\documentclass[letterpaper, 10 pt, conference]{ieeeconf}  

\IEEEoverridecommandlockouts                              

\usepackage{url}
\usepackage{amssymb}

\usepackage{xcolor} 

\usepackage[pdftex]{graphicx}
\DeclareGraphicsExtensions{.pdf,.jpeg,.jpg,.png,.eps}

\usepackage[%
  backend=bibtex      
 ,style=ieee
 ,sorting=none        
 ,sortcites=true      
 ,block=none
 ,indexing=false
 ,citereset=none
 ,isbn=true
 ,url=false
 ,doi=true            
 ,natbib=true         
 ,maxnames=3
 ,minnames=1
]{biblatex}

\usepackage[T1]{fontenc}

\newcommand{\resec}[1]{Section~\ref{sec:#1}}
\newcommand{\refig}[1]{Fig.~\ref{fig:#1}}

\newcommand{\flexbt}{\texttt{flexible\_behavior\_trees}}

\renewcommand{\url}[1]{\ #1}

\title{\LARGE \bf
{Flexible Behavior Trees:
In search of the mythical HFSMBTH for Collaborative Autonomy in Robotics}
}

\author{Joshua M. Zutell\textsuperscript{\textdagger}
\and
David C. Conner\textsuperscript{\textdagger}
\and
Philipp Schillinger\textsuperscript{\textdaggerdbl}%
\thanks{\textsuperscript{\textdagger} Capable Humanitarian Robotics and Intelligent Systems Lab (CHRISLab),%
Department of Physics, Computer Science and Engineering,
Christopher Newport University,
Newport News, VA 23606
{\tt\small \{joshua.zutell.18, david.conner\}@cnu.edu}}
\thanks{\textsuperscript{\textdaggerdbl} Bosch Center for Artificial Intelligence,
Robert-Bosch-Campus 1,
71272 Renningen, Germany,
{\tt\small philipp.schillinger@de.bosch.com}
}
}

\begin{document}

\maketitle
\thispagestyle{empty}
\pagestyle{empty}

\begin{abstract}

In recent years, the model of computation known as Behavior Trees (BT), first developed
in the video game industry, has become more popular in the robotics community for
defining discrete behavior switching.
BTs are threatening to supplant the venerable Hierarchical Finite State Machine (HFSM) model.
In this paper we contrast BT and HFSM, pointing out some potential issues with
the BT form, and advocate for a hybrid model of computation that uses both BT and HFSM
in ways that leverage their individual strengths.  The work introduces a new open-source
package for ROS~2 that extends the Flexible Behavior Engine (FlexBE) to enable
interaction with BT-based behaviors within a HFSM in a way that supports collaborative
autonomy.  Simulation and hardware demonstrations illustrate the concepts.

\end{abstract}

\section{INTRODUCTION}
\label{sec:intro}

This paper advocates for a hybrid model of computation that combines the
strengths of the newer Behavior Trees (BT) with the venerable Hierarchical Finite State
Machine (HFSM) model.
As BTs are the newer model, and have received a great deal of attention
in recent years~\cite{towards_bt_14, BT_Modularize_17, extended_bt_17,
bt_robotics_18, learning_bt_19, blended_bt_19, bt_converge_20, bt_autonomous_nav_20},
this paper is partly a defense of HFSM, but only with the
aim of encouraging the proper use of both BT and HFSM according to their relative strengths.
This paper takes inspiration for its title from a 2017 Game Developers Conference
talk by Bobby Anguelov,
in which he discusses similar ideas in the realm of video games, and  advocates
somewhat humorously for ``the mythical HFSM BT hybrid (HFSMBTH)''~\cite{hfsmbth_17}.

\begin{figure}[!t]
\centering
\includegraphics[width=0.95\linewidth]{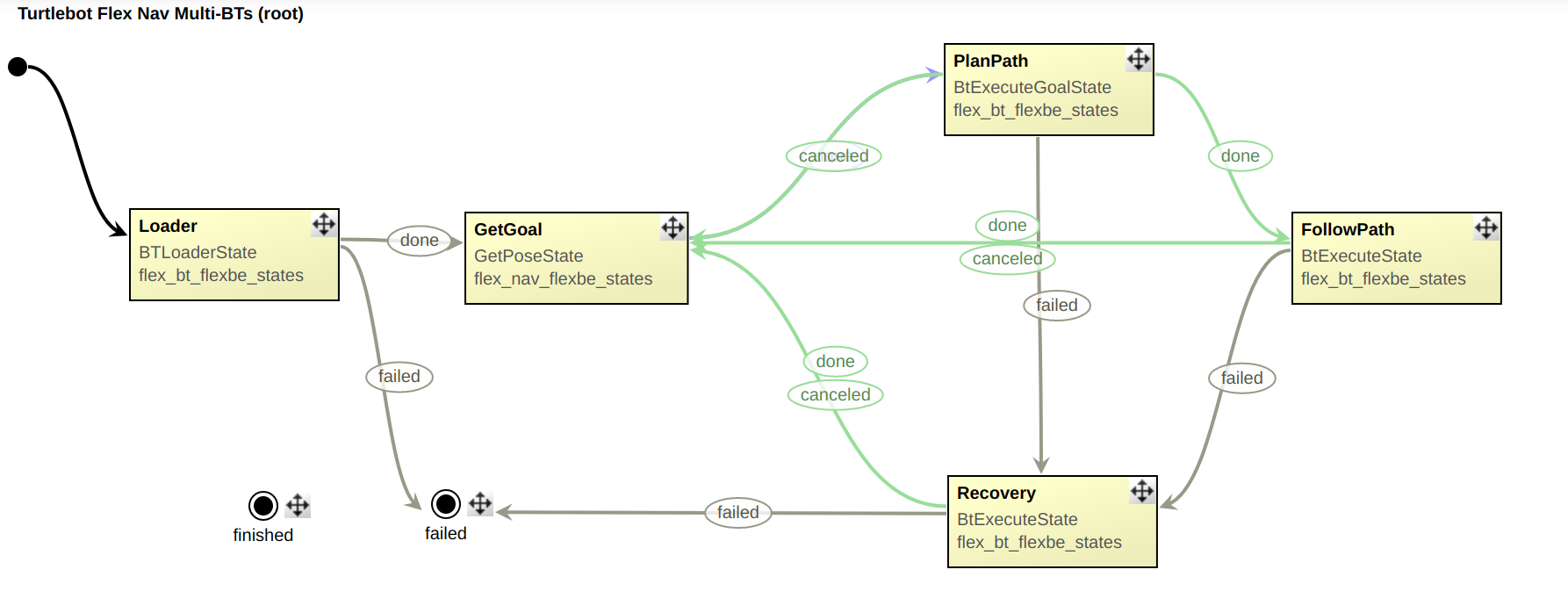}

\vspace{0.05in}

\includegraphics[width=0.95\linewidth]{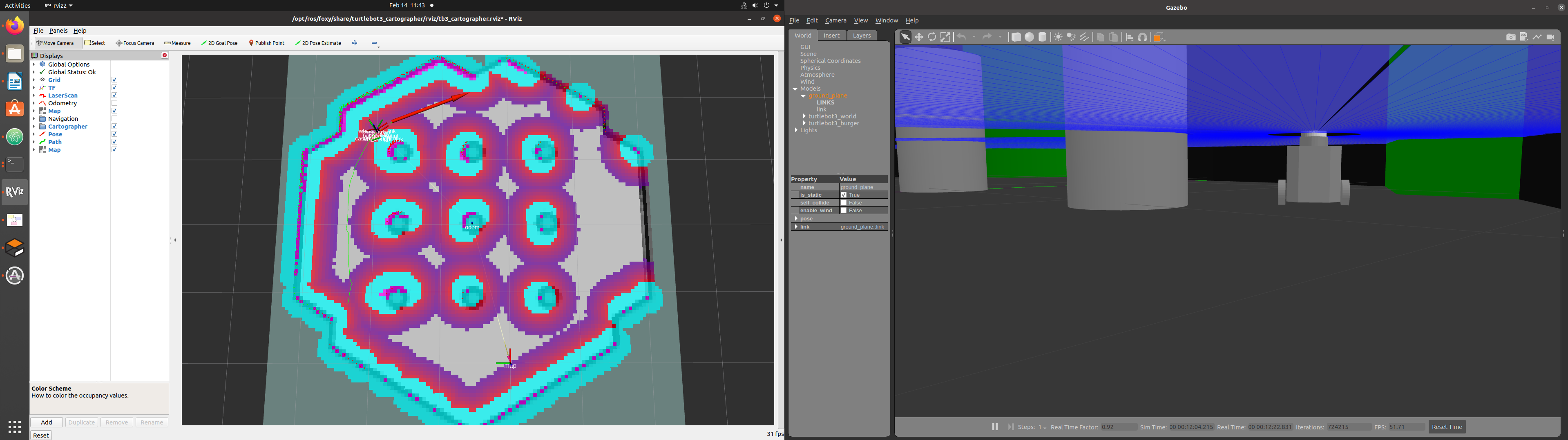}

\vspace{0.05in}

\includegraphics[width=0.95\linewidth]{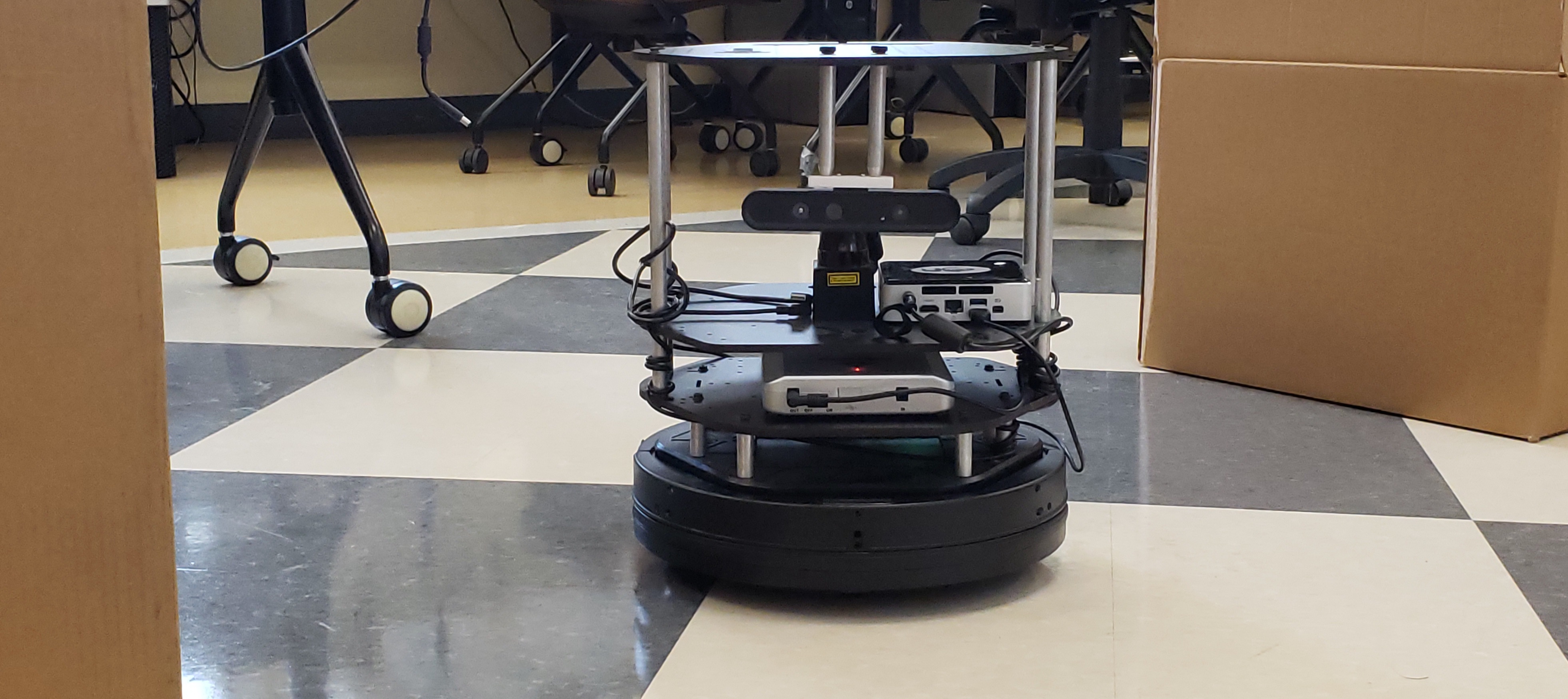}

\caption{Flexible Behavior Trees demonstration (top): FlexBE states that supervise BT execution,
(middle) - demonstration with Turtlebot3 in Gazebo simulation, (bottom) - demonstration on Turtlebot2 hardware.
\label{fig:flexbt_demo}}
\end{figure}

After comparing the BT and HFSM models,
and discussing some potential issues,
we introduce our open-source ROS~2 package for enabling a HFSMBT hybrid within the
existing open-source Flexible Behavior Engine (FlexBE\footnote{\url{http://flexbe.github.io/}}).
The package, dubbed \flexbt{}, allows the user to
incorporate the execution and supervision of a BT within a HFSM,
while preserving the concept of \emph{Collaborative Autonomy} that the authors of
FlexBE describe in~\cite{FlexBE_ICRA_16, ViGIR_JFR_16}.
\refig{flexbt_demo} shows the results of running a demonstration of the FlexBE-based HFSMBT hybrid.

\resec{related} provides an overview of recent research in applications of BTs
to robotics, including a number of open-source BT packages, as well as an overview of
a specific HFSM behavior engine called FlexBE~\cite{FlexBE_ICRA_16, ViGIR_JFR_16}.
\resec{compare} compares BT with simple Finite State Machine (FSM) and
Hierarchical FSM, and highlights some potential issues that favor HFSM.
\resec{flexbt} describes a new ROS~2 package
that allows a user to incoporate FlexBE state implementations that control
execution of a BT in a separate process.
\resec{demo} presents simulation and hardware demonstrations of robot control using
an example HFSMBT hybrid.
\resec{concl} summarizes the contributions.

\section{RELATED WORK}
\label{sec:related}

\subsection{Behavior Trees}

Behavior Trees were first developed in the computer gaming industry as a way
to increase code reusability, incremental design of functionality, and
efficiently test functionality~\cite{BT_Modularize_17, bt_robotics_18}.
In particular, BTs are a model of computation used to define control structures
for in-game non-player characters (NPC). These NPC are hybrid dynamical
systems (HDS), which have both a continuous part like movement and discrete part like decision
making; the BT models this by switching between different controllers.
As such BTs are a middle layer between high-level AI planning systems and low-level
continuous controllers~\cite{towards_bt_14}.

Moreover, as a general model of computation BTs offer the benefits of
combining the functionalities of sequential
behavior compositions, the subsumption architecture, and decision trees to form
robust controllers~\cite{BT_Modularize_17, bt_robotics_18}.
Having the same capabilities as decision trees, BTs have the potential to be used
in machine learning applications~\cite{bt_robotics_18, learning_bt_19}.

To provide the benefits of these vastly different structures, the main structure
 of a basic BT is a directed acyclic graph (i.e. a \emph{tree})~\cite{BT_Modularize_17, bt_robotics_18}.
BTs start at the root node which
periodically emits signals called ticks that are sent depth first down the
highest priority branch (the leftmost branch by graphical convention).
Once a child node receives a tick, it transitions from being idle to executing,
and the tick propagates down to its children, until a subsequent node returns either
 \emph{Success}, \emph{Failure}, or
\emph{Running}. Leaf nodes are either \emph{Actions} or \emph{Conditions}.
\emph{Actions} may change the world state and
may return \emph{Success}, \emph{Failure}, or
\emph{Running}; \emph{Conditions} only report on world state by returning either
\emph{Success} or \emph{Failure}.
These return values are then propagated back up the BT branch to the parent
nodes according to specific rules defined by the three basic
interior control flow nodes: \emph{Fallback}, \emph{Sequence}, and \emph{Parallel} nodes.

\begin{figure}
    \center\includegraphics[width=.9\linewidth]{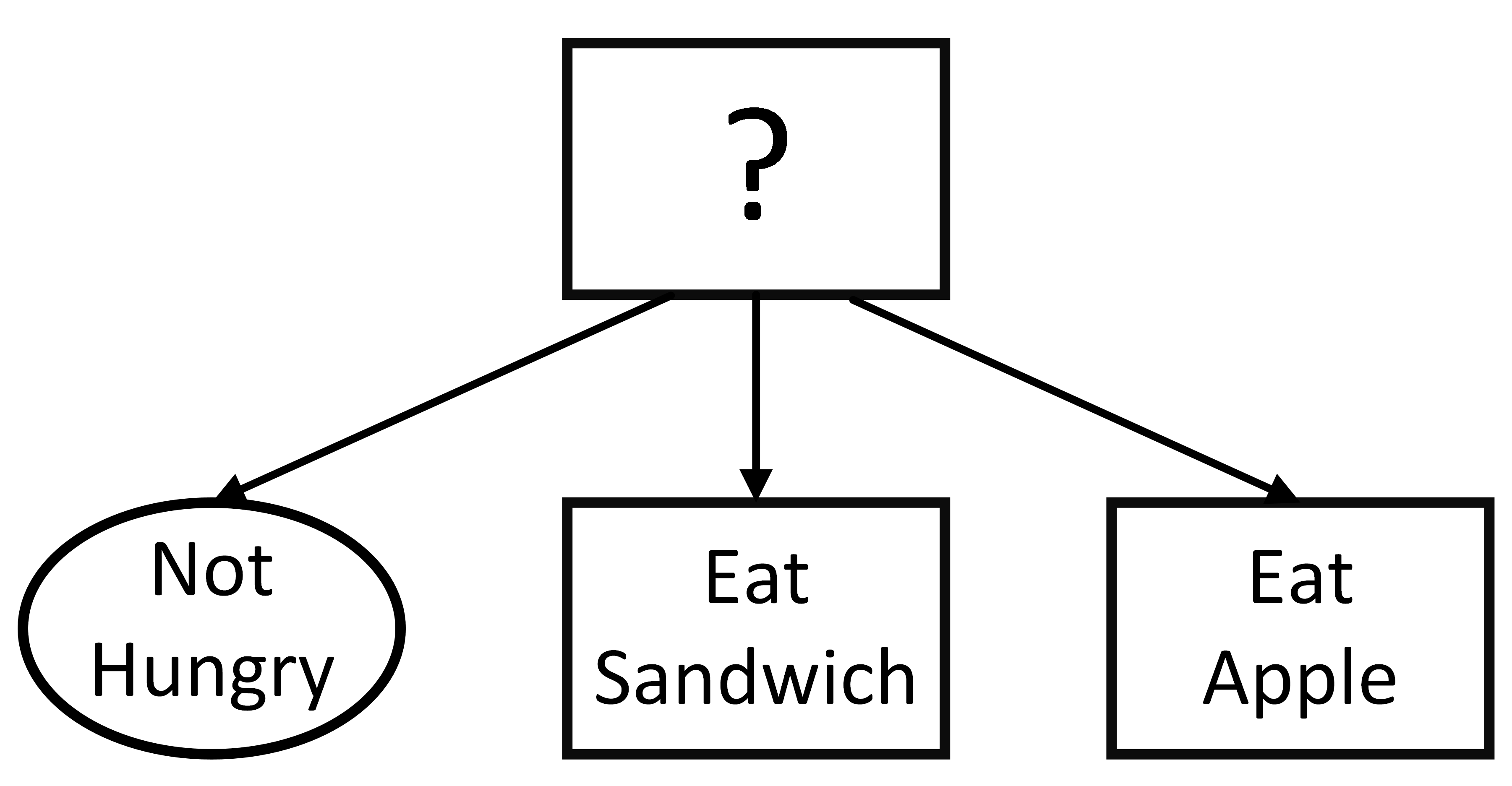}
    \caption{A Fallback node that returns \emph{Success} if ``Not Hungry'' condition returns
    \emph{Success}, or chooses an action that will change world state to "Not Hungry".
    Either ``Eat Sandwich'' or ``Eat Apple'' will satisfy the condition.}
    \label{fig:fallback_node}
\end{figure}

\emph{Fallback} nodes, which are graphically marked as ``\textbf{?}'' in the BT,
are used when there are multiple
ways to achieve a goal as illustrated in \refig{fallback_node}.
Fallback nodes succeed if one of its children succeeds. If the first child of the Fallback
node fails (i.e. the "Not Hungry" condition), then the next child will be ticked to
execute in sequence until a child returns \emph{Success} or \emph{Running}, or all
children have returned \emph{Failure}~\cite{BT_Modularize_17, bt_robotics_18}.

For example, in \refig{fallback_node} if hungry, we will first try
to eat a sandwich.  While this is in process, the \emph{Fallback} node returns \emph{Running}.
On \emph{Successful} completion, the node returns \emph{Success}.  If the
``Eat Sandwich'' action fails (e.g. no sandwich in refrigerator), then the \emph{Fallback} node will tick the ``Eat Apple'' action.
The \emph{Fallback} will only return \emph{Failure} if both ``Eat Sandwich'' and ``Eat Apple'' fail.

\begin{figure}
    \center\includegraphics[width=.9\linewidth]{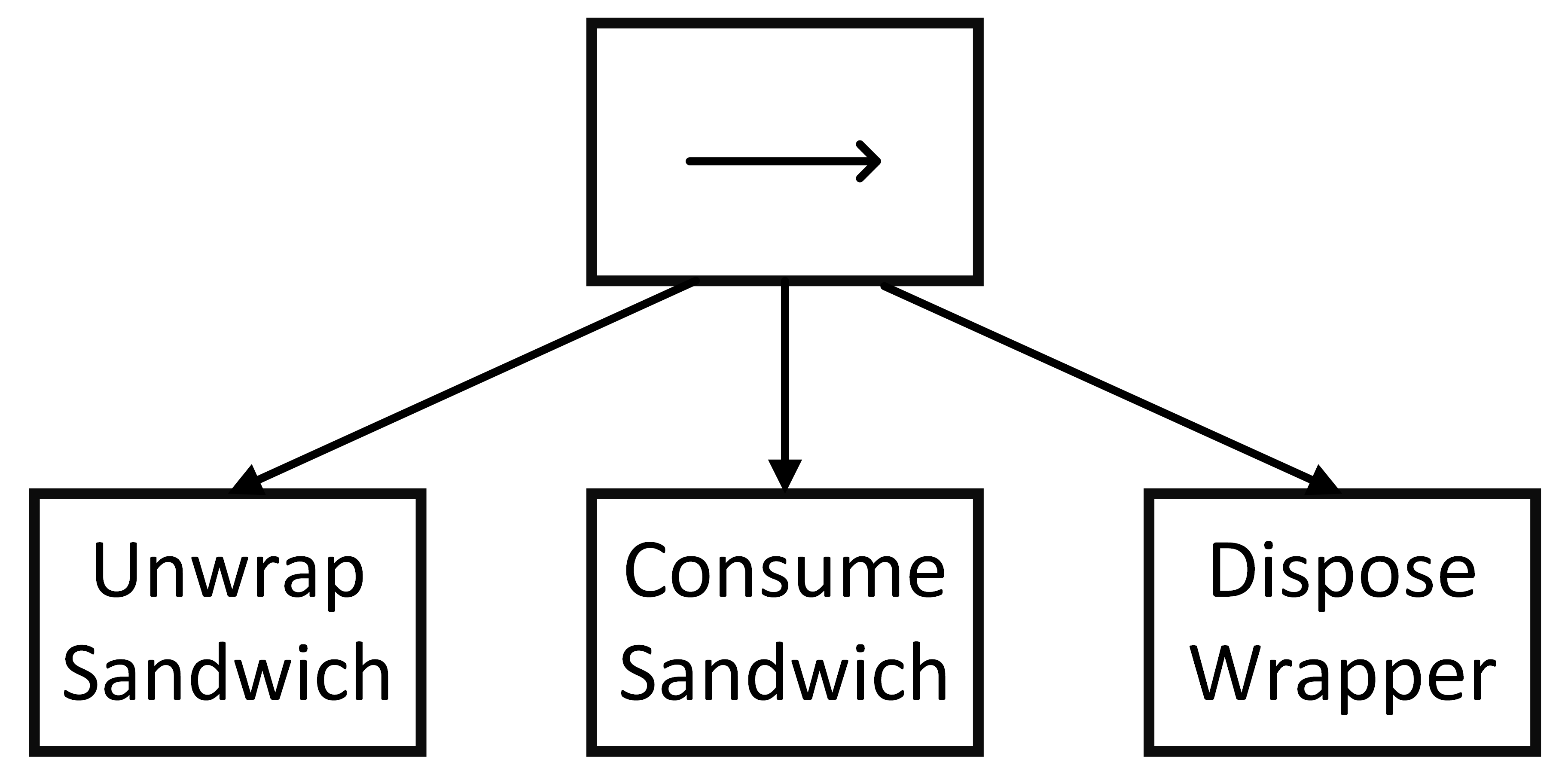}
    \caption{To successfully eat the sandwich we must unwrap it, consume the sandwich,
    and as a good citizen properly dispose of the waste.}
    \label{fig:sequence_node}
\end{figure}

\emph{Sequence} nodes, which are denoted graphically with
``\textbf{\textrightarrow}'', are used when actions need to be completed sequentially
as displayed in \refig{sequence_node}~\cite{BT_Modularize_17, bt_robotics_18}.
Each action is ticked in sequence after the prior
action returns \emph{Success}.
A \emph{Sequence} node returns \emph{Running} or \emph{Failure} as soon as any child
 returns \emph{Running} or \emph{Failure}.
For a \emph{Sequence} to return \emph{Success}, \textbf{all} of
its children must return \emph{Success}.

Lastly, \emph{Parallel} nodes, displayed as ``\textbf{$\rightrightarrows$}'' arrows,  are
used for executing multiple actions simultaneously~\cite{BT_Modularize_17, bt_robotics_18}.
The Parallel node ticks all its children simultaneously. If $M$ children out of
the total number of children $N$ succeed, then the node returns \emph{Success}.
If less than the $M$ children succeed, that is more than $N - M$ children
return \emph{Failure}, then the \emph{Parallel} node returns \emph{Failure}.

Table \ref{table:node_summary} shows the different node types~\cite{BT_Modularize_17, bt_robotics_18}.

\begin{table} [h!]
  \resizebox{\linewidth}{2.75em}{%
	\begin{tabular}{|c|c|c|c|}
	    \hline
	    Node & \emph{Success} & \emph{Failure} & \emph{Running} \\
	    \hline
        	\emph{Action} & Upon completion & Unable to complete & During execution \\
	    \hline
			     \emph{Condition} & If true & If false & Never\\
	    \hline
	        \emph{Fallback} & A child returns \emph{Success} & All children return \emph{Failure} & A child returns \emph{Running}\\
	    \hline
          \emph{Sequence} & All children return \emph{Success} & A child returns \emph{Failure} & A child returns \emph{Running}\\
        \hline
 		      \emph{Parallel} & M or more children return \emph{Success} & More than N - M children return \emph{Failure} & else \\
 		\hline
	\end{tabular}
  }
	\caption{Summary of each type of BT node and return conditions}
    \label{table:node_summary}
	\end{table}

\vspace{-15pt}

The basic BT design says that on each tick of the root node, the BT traverses
the highest priority nodes first.
In theory, this allows the BT to be highly reactive, but requires repeated
checking of conditions.
To avoid large computational burdens of repeated sensor checks,
the concept of a ``belief vector'' that is updated asynchronously
is maintained~\cite{bt_belief_vector_21}.

The common alternative to regular ticking is to use event-driven ticking
where a new tick is generated when an action returns \emph{Success} or \emph{Failure},
but not when an action is \emph{Running}, or ticking when
an external system triggers an event based on a condition change~\cite{bt_misuse_20}.
This approach reduces reactivity of the BT, and ``complexifies'' the BT concept,
requiring additional tools for monitoring events and analyzing trees.  In spite of these
drawbacks, event-driven BTs are said to be the most common and de facto standard
approach in the video game industry~~\cite{bt_misuse_20}.

A powerful concept in designing BTs is the concept of backchaining modular sub-trees from
an overall goal~\cite{bt_robotics_18, blended_bt_19, bt_backward_chaining_19}.
Goal conditions are iteratively expanded
with sub-trees that achieve those conditions.  This concept applies as both an engineering
design approach and algorithm for automated building of a BT.  BTs have also been synthesized
from Linear Temporal Logic (LTL) specifications~\cite{synthesis_bt_17}.

There have been several BT libraries created to implement
BTs in C++, Python, and the gaming framework Unity; examples include the
\texttt{BehaviorTree.CPP}\footnote{\url{https://github.com/BehaviorTree/BehaviorTree.CPP}},
\texttt{py\_trees}\footnote{\url{https://github.com/splintered-reality/py\_trees}}, and
\texttt{NPBehave}\footnote{\url{https://github.com/meniku/NPBehave}}~\cite{bt_implement}.
The C++ BT framework \texttt{BehaviorTree.CPP} is widely used in robotics,
and provides the creation of BTs using XML files with classical nodes and nodes with memory.
To create nodes with memory, each XML file defines input and output ports that are
accessed by a centralized shared memory called blackboard. Using these ports,
the blackboard passes data between different BT nodes where one node’s output data can
become another node’s input~\cite{bt_implement}.
In addition to passing data between nodes, there are decorator nodes such as
a RepeatNode and RetryNode to execute a node multiple times.
The RepeatNode will continue to tick a
child node until the child fails. Meanwhile, a RetryNode will continue to tick a
child node until the child succeeds.
These decorator nodes allow for creating cyclic or partially cyclic behaviors to compensate
for an inherently acyclic data structure.

BTs have also made their way into the popular Robot Operating System (ROS),
maintained by Open Robotics\footnote{\url{https://www.openrobotics.org/company}}
~\cite{quigley_ros_2009}.
The Move Base Flex navigation system uses Behavior Trees instead of HFSM as a replacement
for ROS~1's \texttt{navigation} package~\cite{movebaseflex}.
With the introduction of ROS~2, the
\texttt{navigation2}\footnote{\url{https://github.com/ros-planning/navigation2}} package
uses behavior trees for task orchestration~\cite{navigation2}.  It should be noted
that both have implicit FSM in their handling of receipt of goals via ROS messages,
and transition to execution of the BT.

\subsection{HFSM and the Flexible Behavior Engine}

Prior to the advent of BT, many modern robot systems used Hierarchical Finite State Machines (HFSM)
to define a high-level system behavior used to coordinate between subsystems~\cite{FlexBE_ICRA_16}.
An early ROS~1 package called SMACH\footnote{\url{http://wiki.ros.org/smach}} (State MACHine)
allowed users to define such HFSM and execute them on a robot.

In the DARPA Robotics Challenge (2012-15),
it was desired for the robot to be a member of a team and not purely autonomous;
thus, there was a need for a supervisor to preempt behaviors and reconfigure the robot behaviors
in response to changing conditions encountered during a disaster response~\cite{ViGIR_JFR_16}.
This ability to interact with the robot system was dubbed \emph{Collaborative Autonomy}~\cite{ViGIR_JFR_16}.
As SMACH was not designed for collaborative autonomy,
Team ViGIR developed the Flexible Behavior Engine (FlexBE) as a major extension to SMACH.
FlexBE supports adjustable autonomy, preemptive state transitions, and online
adjustments to behaviors that support collaborative autonomy.
FlexBE was released as an open-source ROS package~\cite{FlexBE_ICRA_16}.

\begin{figure}
    \center\includegraphics[width=.99\linewidth]{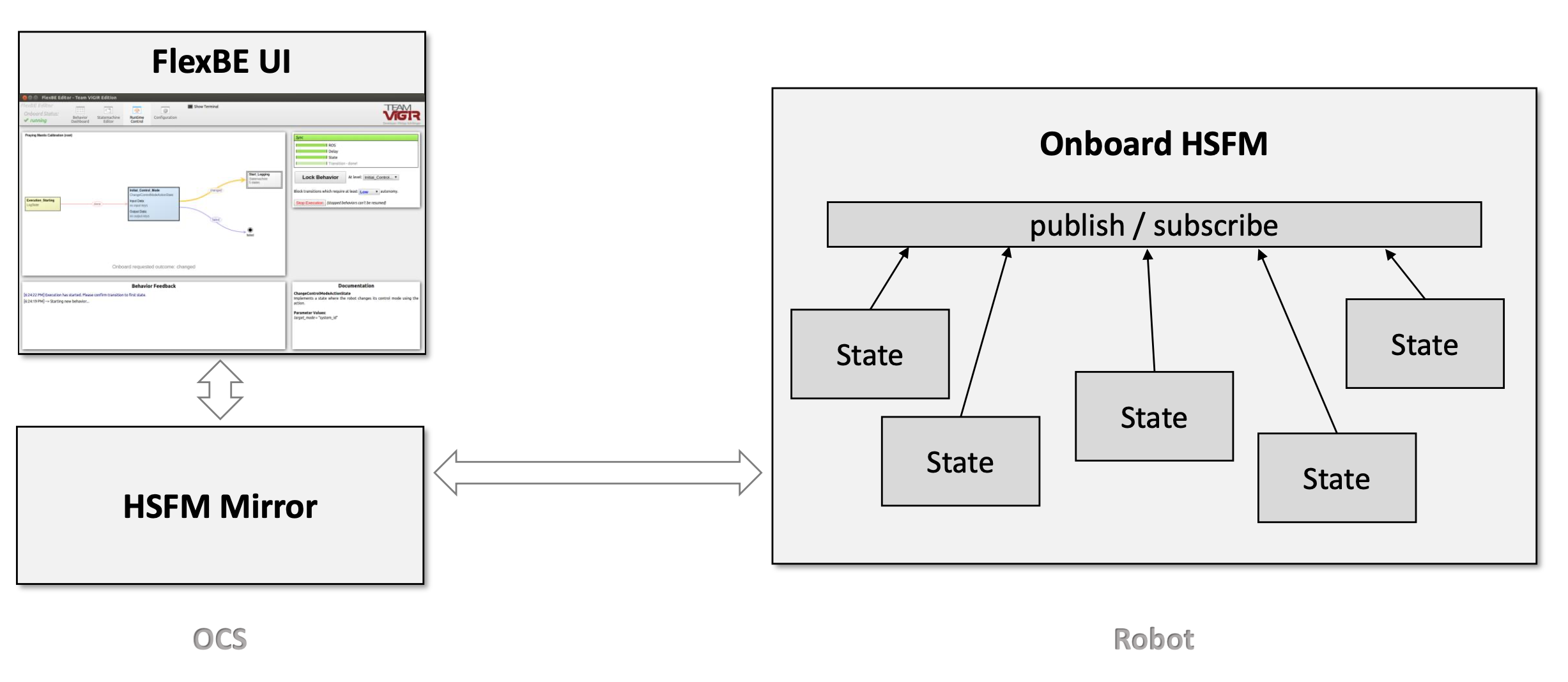}
    \caption{FlexBE system architecture (from~\cite{zutell_22}).}
    \label{fig:flexbe_arch}
\end{figure}

\refig{flexbe_arch} shows the four parts of the FlexBE system:
\begin{itemize}
\item Onboard (robot) Behavior Executive (OBE)
\item OCS User Interface (FlexBE UI)
\item OCS HFSM ``mirror''
\item Various Python-based state implementations
\end{itemize}
On a robot onboard computer, the Onboard Behavior Executive coordinates execution of the
Python-based state implementations and handles coordination of the publish/subscribe
to various topics ~\cite{FlexBE_ICRA_16, ViGIR_JFR_16}. On a separate computer, FlexBE provides
a desktop graphical user interface for the OCS that enables easier development of HFSMs and
for monitoring their execution in realtime. The OCS mirror node acts as a bridge between the FlexBE UI
and OBE by ``mirroring'' the status of the onboard state machine without actually executing
the states. The mirror communicates with the OBE using standard ROS messaging, and with
the FlexBE UI using a JavaScript ROS messaging wrapper. Lastly, the Python-based state implementations
encode the HFSM state nodes, and are \emph{actions} that interact with the robot system
processes (e.g. path planning, sensor processing, navigation control). Like SMACH, FlexBE
enables the passing of \emph{user data} from one state to the next. This paper builds upon
recent work that has upgraded FlexBE to work with ROS~2~\cite{zutell_22}.

FSM can be defined using correct-by-construction formal methods techniques,
including synthesis of reactive automata from LTL specifications~\cite{waldo_17}.
These techniques have been applied to generating a realized FlexBE-specific HFSM for
direct execution of robot control~\cite{atlas_synthesis_16}.

In this paper we introduce a new ROS~2 package that enables a user to execute a BT
from within a FlexBE HFSM.  This realizes the HFSMBTH concept, and enables
modular BT sub-trees to be directly incorporated into a reactive HFSM that
enables collaborative autonomy.

\section{BT AND HFSM COMPARISON}
\label{sec:compare}

In this section we discuss the advantages and limitations of both BTs and HFSMs.
Following~\cite{hfsmbth_17}, we advocate
for a hybrid combination of BT and HFSM that is greater than the sum of their parts.
We address three specific claims for benefits of BTs -- modularity, two-way data flow,
and lack of hidden state -- along with their limitation of being acyclic.

While BTs and HFSMs are both equivalent in their computing power, one of the claimed benefits
of BTs is their modularity given the weak interdependence
between nodes due to the simplified \emph{Success}, \emph{Failure}, and \emph{Running}
return values~\cite{bt_robotics_18, bt_backward_chaining_19}. One can substitute a
sub-tree that achieves a pre-condition, or expand an action with a
BT that achieves the desired outcome.  It is clear however, that this modularity
is not absolute and is not free from unintended or emergent impacts on overall system
behavior.

Consider the simple examples from \refig{fallback_node}
and~\ref{fig:sequence_node}.
Using backchaining, the sequence node subtree in~\refig{sequence_node}
achieves the ``Eat Sandwich'' action required in~\refig{fallback_node}. Under
normal circumstances , this will satisfy the \emph{Fallback} node and achieve the
condition ``Not Hungry.'' However, consider a case when disposal of the sandwich
wrapper fails because the garbage can is full.

In this case, the ``Eat Sandwich'' action fails, and the \emph{Fallback} ticks ``Eat Apple,''
which leads to a case of overeating that was not the desired behavior.  One could
add a ``Blackboard'' condition that recognizes the sandwich was already consumed,
but this introduces a hidden state into the BT.  A good designer might recognize
this pitfall, but they do not have control over how the subtrees may be used in
the future.  It is not always possible to know the best design a priori.  One could
backchain an action that empties the trash can if it is full before attempting to dispose
of wrapper, but that introduces what could be a relatively low priority task that preempts
continuation of an actually higher priority task. We are not arguing that a good BT design
cannot overcome this example, only that the modularity is not absolute. Solutions
may require ad hoc engineering that violate the basic computational model of the BT,
and result in the introduction of hidden control state into the system.

In contrast, a comparable HFSM design would treat the sequence node in \refig{sequence_node}
as a simple FSM with multiple labeled outcomes. The overall \emph{Success} would be the
primary outcome, with potential \emph{Failure} on the separate child actions with unique labels.
Making these additional connections is additional work, but the choice becomes explicit.
Furthermore, the potential transitions are amenable to formal analysis and correct-by-construction
synthesis~\cite{atlas_synthesis_16}.

In addition to modularity, BTs are considered favorable over HFSMs due to BTs having two way
transitions while HFSMs only have one way transitions~\cite{BT_Modularize_17, bt_robotics_18}.
BTs are able to transition up and down the tree using function calls while HFSMs can only
transition to the next state sequentially analogous to a \texttt{Goto} statement~\cite{bt_robotics_18}.
Although HFSMs only have one way transitions, HFSMs in both SMACH and FlexBE allow for user
data to be defined that is passed along from state to state. While not two-way data flow in
the sense that BT have, this concept increases the adaptability of the FSM.  As illustrated
above, the basic two-way data flow in BTs does not address some conditions without requiring
shared data and the introduction of hidden state.

The BT is fundamentally an acyclic directed graph~\cite{bt_misuse_20}.
This has two implications that we highlight here.  First, this structure
imposes a total order on the resulting behavioral system with
an implicit priority that is defined by the pre-conditions.  Functionally, the
BT is equivalent to a large collection of ordered \texttt{if-elif-else} blocks~\cite{bt_misuse_20}.
While the priority total ordering is implicit in the structure, it is not readily apparent
for larger trees~\cite{bt_misuse_20}.

The second significant implication is that BTs are often using a fundamentally acyclic data
structure to control fundamentally cyclic behaviors~\cite{ hfsmbth_17, bt_misuse_20}.
In contrast, defining looping behaviors within an HFSM is natural and readily apparent from
the graphical structure of the HFSM.  Generating cyclical behaviors using a BT
paradigm is clearly possible, but does so via external triggers and decorators;
the cyclical behavior is not obvious from the graphical structure of the BT.

Finally, the FlexBE system supports
collaborative autonomy~\cite{FlexBE_ICRA_16, ViGIR_JFR_16}.
HFSM are a natural representation for humans to follow, and the transitions readily
encode the expected transitions and contingency plans in the form of a script~\cite{ViGIR_JFR_16}.
FlexBE HFSM support adjustable autonomy and operator interactions via preemption and
blocked transitions.

The remainder of this paper describes a system for incorporating BT subtrees into
a FlexBE HFSM to support cyclic behaviors and the best use of BTs
according to their relative strengths.  The work enables the
``mythical HFSMBT hybrid'' that builds upon the strengths of both BT and HFSM to
achieve something greater than either alone~\cite{hfsmbth_17}.

\section{FLEXIBLE BEHAVIOR TREES}
\label{sec:flexbt}

To combine the benefits of HFSMs and BTs, we developed the
\flexbt{}\footnote{\url{https://github.com/FlexBE/flexible\_behavior\_trees}} ROS package
 to enable embedding BTs into FlexBE HFSMs.
The package contains a BT server node derived from the ROS~2 \texttt{navigation2}
\texttt{nav2\_behavior\_tree} package, which is based on the \texttt{BehaviorTree.CPP} framework.
Following the ROS~2 \texttt{navigation2} BT server format, the BTs are encoded as XML files,
which are parsed and loaded into the BT server where they are stored by behavior name.
The \flexbt{} package defines custom ROS Actions --  \texttt{BtLoad} and
\texttt{BtExecute} -- and corresponding FlexBE state implementations
\texttt{BtLoadState}, \texttt{BtExecuteState}, and \texttt{BtExecuteGoalState}.

FlexBE initiates BT execution by sending a \texttt{BtExecute} action goal
to the BT server node with a designated behavior name to execute.
To monitor progress of executing a BT, the BT
server sends \texttt{BtExecute} action feedback messages containing which BT nodes are
active, current location of the robot, and how long the BT has been executing.
Once a BT has finished executing, the BT server will respond to the
\texttt{BtExecute} result message that indicated whether the BT returned
\emph{Success}, \emph{Failure}, or if execution was canceled due to a \texttt{BtExecute}
cancel messsage.

The \flexbt{} package also defines three
states: \texttt{BtLoaderState}, \texttt{BtExecuteState}, and \texttt{BtExecuteGoalState}
that interface with the BT server node.
The \texttt{BtLoaderState} creates and sends a custom \texttt{BtLoad} goal message
containing a list of BT XML files for the BT server to load.
The motivation for this \texttt{BtLoaderState} is to allow a user to load
relevant BTs prior to execution. Next, the \texttt{BtExecuteState}
sends a goal message to the BT server to execute a specified BT by name.
The \texttt{BtExecuteGoalState} works similarly, but adds either one or multiple \texttt{PoseStamped}
message to the \texttt{BtExcute} goal for
basic navigation behaviors and other behaviors requiring user defined goals.
After sending the goal to the BT server and receiving the result back from the BT server,
the FlexBE BT state implementations return transitions labeled \texttt{done}, \texttt{failed},
or \texttt{canceled} corresponding to the BT results.

Using FlexBE and \flexbt{} states, larger BTs can be deconstructed
into smaller sub-trees, while using FlexBe’s
collaborative autonomy benefits by adding more FlexBE states to the HFSM.
This allows more user intervention while preserving the modularity benefits of BTs.
These sub-trees on the BT server are able to share data through
\texttt{BehaviorTree.CPP}’s blackboard.

\section{DEMONSTRATION}
\label{sec:demo}

\begin{figure}
    \center\includegraphics[width=.65\linewidth]{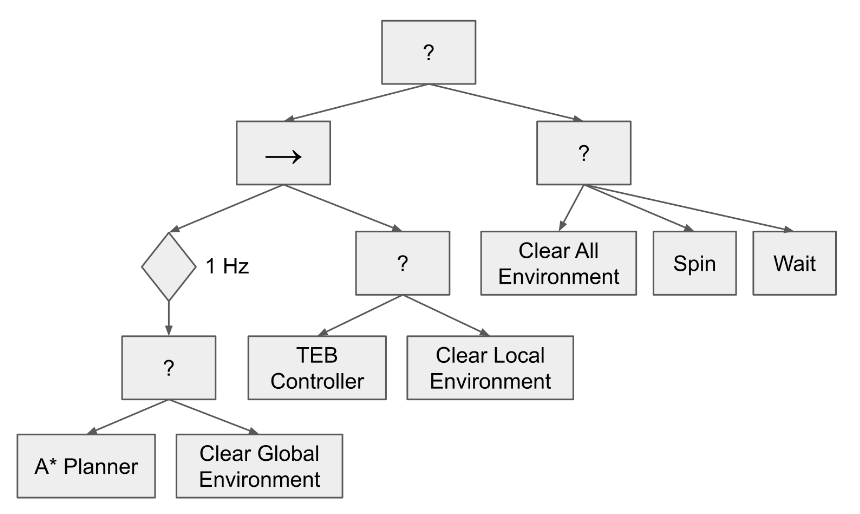}
    \caption{ROS~2 Navigation2 Stack example navigation BT from~\cite{navigation2}.}
    \label{fig:nav2_bt}
\end{figure}

\begin{figure}
    \center\includegraphics[width=.9\linewidth]{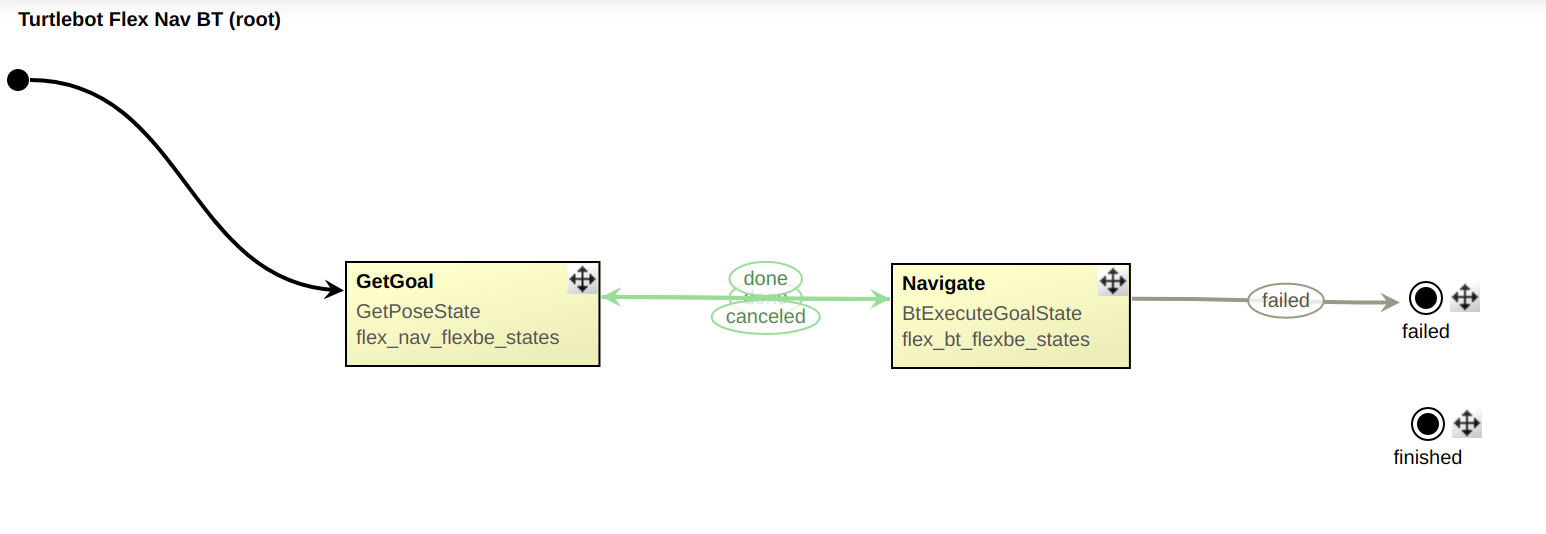}
    \caption{Simple navigation HFSMBTH behavior in FlexBE.}
    \label{fig:simple_behavior}
\end{figure}

In this section we will discuss two demonstrations of the Flexible Behavior Trees.
The first demonstration loaded a BT similar to the
ROS~2 \texttt{navigation2} BT in \refig{nav2_bt} and executed from a FlexBE HFSM shown
in \refig{simple_behavior}; this makes the FSM that is implicit in the
\texttt{navigation2} explict to support collaborative autonomy.
Meanwhile, the second demonstration split the BT from \refig{nav2_bt} into separate sub-trees -- global
planner, controller, and recovery behaviors -- interfaced by separate FlexBE
state implementations shown in \refig{flexbt_demo}.
The complete setup for both the simulation and hardware demonstrations
are provided\footnote{\url{https://github.com/FlexBE/flexbt\_turtlebot3\_demo}}.

\vspace{-5pt}

In the first demonstration, the HFSMBTH behavior in \refig{simple_behavior} gets a
goal from the user and sends it to
the BT server to navigate independently to the goal. This is similar to operation
of the ROS~2 Navigation2 Stack with additional collaborative autonomy that
 allows the user to cancel and re-enter another goal if they do not like the
 selected target.
After navigating to the goal, the HFSMBTH loops to ask the user for another goal.
The HFSMBTH returns \emph{failed} if the BT result returns failure.

\begin{figure}
    \center\includegraphics[width=\linewidth]{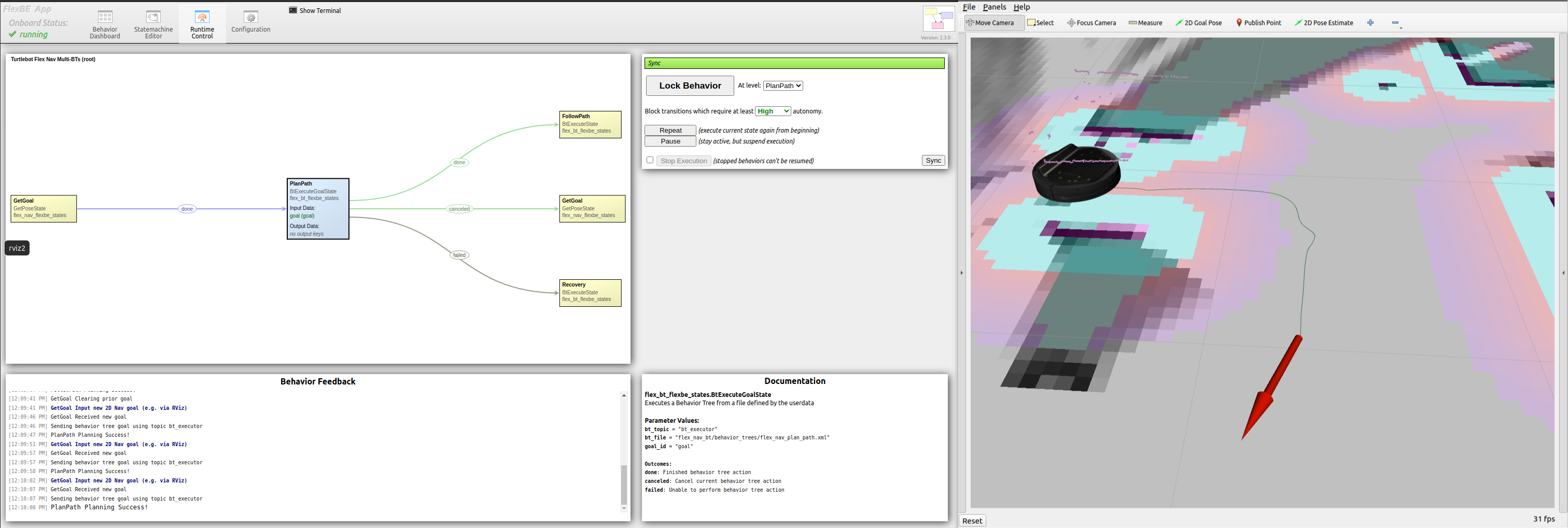}
    \caption{During the hardware demonstration of the \refig{flexbt_demo} behavior,
    the FlexBE UI is set to require
    high autonomy level transitions.  After planning is successful, the FlexBE behavior
    pauses and requires the
    user to confirm acceptance of the plan to the designated goal that is displayed in the
    RViz UI.
    In this collaborative autonomy mode, the user
    can select \emph{done} to allow navigation, or \emph{canceled} to return to \emph{GetGoal}
    state to allow input of another goal to re-trigger the planning phase, or
    \emph{failed} to trigger a recovery behavior.}

    \label{fig:flexbe_hardware_demo}
\end{figure}

The second demonstrated HFSMBTH behavior, as shown in \refig{flexbt_demo}, splits the first
demonstration BT into sub-trees providing for more user intervention.
The behavior first loads all the BTs that will be executed and then asks the user
for a target goal. After receiving this goal, the HFSMBTH behavior transitions
into creating a global plan to the goal. Using the adjustable autonomy feature of FlexBE,
the user can first view the returned path
(using RViz in this example),
and either accept the plan as is by allowing the HFSM  transition,
or choose the \emph{canceled} outcome to require input of another
target goal as illustrated in \refig{flexbe_hardware_demo}. Upon accepting the
plan, the HFSMBTH transitions into navigating towards the goal using a controller.
The BT action controller is able to receive the plan from the global
planner through the \texttt{BehaviorTree.CPP}’s blackboard.
During navigation, the user is able to monitor the execution and preempt the
navigation state via the FlexBE UI. Moreover, both the global planner and navigation states
allow the user to manually stop execution or force transition into performing
recovery behaviors in case of an emergency. The recovery behaviors state will perform
the specified recovery actions in the BT, and upon successful completion transitions to
ask the user for another goal. In this simple example, if the recovery behaviors
return \emph{Failure}, the state return \emph{failed} transition and the
HFSMBTH ends navigation.

These HFSMBTH behaviors were tested in Gazebo-based simulation using a TurtleBot3 model,
and on hardware with a TurtleBot2 as shown in \refig{flexbt_demo}.
The software runs under Ubuntu 20.04 using the ROS~2 ``Foxy Fitzroy'' distribution.
The demonstrations use the ROS~2 versions of FlexBE, Gazebo, RViz2, and
\texttt{navigation2} software, in addition to standard sensor drivers.
Complete directions are provided.

\section{CONCLUSIONS}
\label{sec:concl}

This paper advocates for a hybrid model of computation that combines the
strengths of the newer Behavior Trees (BT) with the venerable Hierarchical Finite State
Machine (HFSM) model.  After discussing the relative strengths and weaknesses of
BT and HFSM, we introduce a new open-source ROS~2 package
\flexbt{}\footnote{\url{https://github.com/FlexBE/flexible\_behavior\_trees}}
that enables a user to provide supervisory control of a behavior tree from within
a FlexBE state machine.  This enables collaborative autonomy, while leveraging the
relative strengths of both BT and HFSM as the appropriate level of abstraction.
Simulation and hardware demonstrations are presented and released
open-source.


\begin{small}
\printbibliography
\end{small}

\end{document}